\documentclass[10pt,twocolumn,letterpaper]{article}

\usepackage{cvpr}
\usepackage{times}
\usepackage{epsfig}
\usepackage{graphicx}
\usepackage{amsmath}
\usepackage{amssymb}

\usepackage{url}
\usepackage{wrapfig, rotating}
\usepackage{algorithmic, algorithm}
\usepackage{amsthm}
\usepackage[mathscr]{eucal}
\usepackage{amsbsy}
\usepackage{bm}
\usepackage{paralist}
\usepackage{xspace}
\usepackage{caption,cite, setspace}

\makeatletter
\newsavebox\myboxA
\newsavebox\myboxB
\newlength\mylenA

\newcommand*\xbar[2][0.75]{%
	\sbox{\myboxA}{$\m@th#2$}%
	\setbox\myboxB\null
	\ht\myboxB=\ht\myboxA%
	\dp\myboxB=\dp\myboxA%
	\wd\myboxB=#1\wd\myboxA
	\sbox\myboxB{$\m@th\overline{\copy\myboxB}$}
	\setlength\mylenA{\the\wd\myboxA}
	\addtolength\mylenA{-\the\wd\myboxB}%
	\ifdim\wd\myboxB<\wd\myboxA%
	\rlap{\hskip 0.5\mylenA\usebox\myboxB}{\usebox\myboxA}%
	\else
	\hskip -0.5\mylenA\rlap{\usebox\myboxA}{\hskip 0.5\mylenA\usebox\myboxB}%
	\fi}
\makeatother

\newcommand{\Sec}[1]{\hyperref[sec:#1]{\S\ref*{sec:#1}}} 
\newcommand{\Eqn}[1]{\hyperref[eq:#1]{(\ref*{eq:#1})}} 
\newcommand{\Fig}[1]{\hyperref[fig:#1]{Figure~\ref*{fig:#1}}} 
\newcommand{\Tab}[1]{\hyperref[tab:#1]{Table~\ref*{tab:#1}}} 
\newcommand{\Thm}[1]{\hyperref[thm:#1]{Theorem~\ref*{thm:#1}}} 
\newcommand{\Lem}[1]{\hyperref[lem:#1]{Lemma~\ref*{lem:#1}}} 
\newcommand{\Prop}[1]{\hyperref[prop:#1]{Property~\ref*{prop:#1}}} 
\newcommand{\Cor}[1]{\hyperref[cor:#1]{Corollary~\ref*{cor:#1}}} 
\newcommand{\Def}[1]{\hyperref[def:#1]{Definition~\ref*{def:#1}}} 
\newcommand{\Alg}[1]{\hyperref[alg:#1]{Algorithm~\ref*{alg:#1}}} 
\newcommand{\Ex}[1]{\hyperref[ex:#1]{Example~\ref*{ex:#1}}} 

\newcommand{\Tra}{^{\rm T}} 

\newcommand{\M}[1]{{\bm{\mathbf{\MakeUppercase{#1}}}}} 

\newcommand{\T}[1]{\boldsymbol{\mathscr{\MakeUppercase{#1}}}} 

\newcommand{\TA}{\T{A}}

\newcommand{\TS}{\T{S}}

\newcommand{\TU}{\T{U}}
\newcommand{\TV}{\T{V}}

\newcommand{\fft}{ \mbox{\tt fft} }
\newcommand{\ifft}{ \mbox{\tt ifft} }

\usepackage{graphics}
\usepackage{algorithmic,algorithm}
\usepackage{amsthm}                 
\usepackage[mathscr]{eucal}         
\usepackage{amsbsy}                 
\usepackage{bm}                     
\usepackage{paralist}               
\usepackage{xspace}                 

\newcommand{\barr}{\left[ \begin{array} }
	\newcommand{\earr}{ \end{array} \right] }
\newcommand{\ars}[1]{\left[ \begin{array}{#1}}
	\newcommand{\are}{\end{array} \right] }
\newcommand{\oars}[1]{\begin{array}{#1}}
	\newcommand{\oare}{\end{array}}
\newcommand{\eqs}{\begin{eqnarray}}
\newcommand{\eqe}{\end{eqnarray}}
\newcommand{\eqsn}{\begin{eqnarray*}}
	\newcommand{\eqen}{\end{eqnarray*}}

\newcommand{\ens}{\begin{enumerate}}
	\newcommand{\ene}{\end{enumerate}}
\newcommand{\its}{\begin{itemize}}
	\newcommand{\ite}{\end{itemize}}
\newcommand{\des}{\begin{description}}
	\newcommand{\dee}{\end{description}}

\newtheorem{theorem}{Theorem}[section]
\numberwithin{theorem}{subsection}

\numberwithin{lemma}{subsection}
\newtheorem{defn}{\indent \bf Definition}[section]
\numberwithin{defn}{subsection}

\numberwithin{conjecture}{subsection}
\newtheorem{remark}{Remark}[section]
\numberwithin{remark}{subsection}

\usepackage[pagebackref=true,breaklinks=true,letterpaper=true,colorlinks,bookmarks=false]{hyperref}

\cvprfinalcopy 

\def\cvprPaperID{2230} 

\ifcvprfinal\fi
\begin{document}

\title{Denoising and Completion of 3D Data via Multidimensional dictionary learning}

\author{Zemin Zhang, Shuchin Aeron\\
Department of EECS, Tufts University\\
Medford, MA 02155, USA\\
{\tt\small zemin.zhang@tufts.edu, shuchin@ece.tufts.edu}}

\maketitle

\begin{abstract}
   In this paper a new dictionary learning algorithm for multidimensional data is proposed. Unlike most conventional dictionary learning methods which are derived for dealing with vectors or matrices, our algorithm, named K-TSVD, learns a multidimensional dictionary directly via a novel algebraic approach for tensor factorization as proposed in \cite{Braman2010,KBN,Kilmer2011641}. Using this approach one can define a tensor-SVD and we propose to extend K-SVD algorithm used for 1-D data to a K-TSVD algorithm for handling 2-D and 3-D data. Our algorithm, based on the idea of sparse coding (using group-sparsity over multidimensional coefficient vectors), alternates between estimating a compact representation and dictionary learning. We analyze our K-TSVD algorithm and demonstrate its result on video completion and multispectral image denoising.
\end{abstract}

\section{Introduction}
Sparsity driven signal processing has been widely used in many areas across computer vision and image analysis, such as image restoration and classification \cite{ksvd,superresolution,dlclassification}. The main principle driving the gains is the idea of sparse coding, i.e. the underlying signal is compactly represented by a few large coefficients in the overcomplete dictionary, while the noise and the sampling process are incohrent. 
Since the performance heavily relies on the chosen dictionary, a lot of dictionary learning algorithms are developed to obtain dictionaries that are more adapted to the signal than the predefined ones, such as wavelet and DCT. In \cite{ksvd}, Aharon \emph{et al}. proposed an algorithm called K-SVD, which efficiently learns an overcomplete dictionary from a set of training signals. The method of optimal directions (MOD)\cite{MOD} shares the same effective sparse coding principle for dictionary learning as K-SVD. Discriminative K-SVD algorithm (D-KSVD) proposed in \cite{dksvd} improved the K-SVD method by unifying the dictionary and classifier learning processes. \cite{bomp} efficiently accelerated the K-SVD algorithm and reduced its memory consumption using a batch orthogonal matching pursuit method.

When the signal is not limited to two dimensional signals, traditional methods generally embed the high dimensional data into a vector space by vectorizing the data points; therefore the conventional matrix based approaches can still be used. This kind of vectorization, however, will lead to a poor sparse representation since it breaks the original multidimensional structure of the signal and reduce the reliability of post processing. To this end, some dictionary learning techniques have been explored based on different tensor decompositions such as CP decomposition \cite{kcpd,CP2}, Tukcer Decomposition\cite{tucker1,tucker2,tucker3} and tensor-SVD\cite{tsvddl}. In \cite{kcpd}, the authors developed an algorithm called K-CPD which learns high order dictionaries based on the CP decomposition. \cite{tucker1} proposed a tensor dictionary learning algorithm based on the Tucker model with sparsity constraints over its core tensor, and applied gradient descent algorithm to learn overcomplete dictionaries along each mode of the tensor (see \cite{Tuck1966c} for definition of tensor modes). Peng et al. \cite{tucker2} presented a tensor dictionary learning algorithm based on Tucker model with Group-block-sparsity constraint on the core tensor with good performance. 



In this paper, we present a novel multidimensional dictionary learning approach based on a notion of tensor-SVD proposed in \cite{Braman2010,KBN,Kilmer2011641}. Essentially the t-SVD is based on an operator theoretic interpretation of the 3rd order tensors \cite{Braman2010}, as linear operators over the set of 2-D matrices. This framework has recently been used for dictionary learning for 2-D images in \cite{tsvddl}, but the authors there employ a different algorithm and the problem considered is tomographic image reconstruction. Moreover we will also consider the problem of filling in missing data by sparse coding using the learned dictionary.

The paper is organized as follows. In section 2 we go over the definitions and notations, then illustrate the main differences and advantages over other tensor decomposition methods. Section 3 formulates the objective function for tensor dictionary learning problem using t-SVD, by introducing the ``tubal sparsity" of third-order tensors. Our tensor dictionary learning model and detailed algorithm to solve the problem are presented in Section 4. In Section 5 we show experiment results on third order tensor completion and denoising. Finally we conclude our paper in Section 6.

\section{Brief Overview of T-SVD Framework}
\label{sec:2}
\subsection{Notations and Preliminaries}
In this part we briefly describe the notations used throughout the paper, and the t-SVD structure proposed in \cite{Braman2010,Kilmer2011641,KBN}.

A tensor is a multidimensional array of numbers. For example, vectors are first order tensors, matrices are second order tensors. Tensors of size $n_1 \times n_2 \times n_3$ are called third order tensors. In this paper, third order tensors are represented in bold script font $\T{A}$.

A \emph{\textbf{Slice}} of an $n$-th order tensor is a $2$-D section defined by fixing all but two indices. For a third order tensor $\T{A}$, we will use the Matlab notation $\TA(k, :, :)$ , $\TA(:, k, :)$ and $\TA(:, :, k)$ to denote the $k$-th horizontal, lateral and frontal slices. $\TA^{(k)}$ is particularly used to represent $\TA(:, :, k)$, and $\overrightarrow{\TA}_k$ represents $\TA(:,k,:)$. We also call such $\overrightarrow{\TA}_k$ \textbf{\emph{tensor columns}}.

A \emph{\textbf{Fiber}} (or \emph{\textbf{Tube}}) is a $1$-D section obtained by fixing all indices but one. For a third order tensor, $\TA(:, i, j)$, $\TA(i, :, j)$ and $\TA(i, j, :)$ denote the $(i, j)$-th mode-$1$, mode-$2$ and mode-$3$ fiber. Specifically we let $\vec{a} \in \mathbb{R}^{1 \times 1 \times n_3}$ denote an $n_3$-tube. 

The approach in \cite{Braman2010,KBN,Kilmer2011641} rests on defining a multiplication operation, referred to as the tensor-product (t-product) between two third order tensors. This is done by using a commutative operation, in particular circular convolution between tensor tubes as defined below.

\begin{defn} \textbf{(t-product)} The t-product between $\T{A} \in \mathbb{R}^{n_1 \times n_2 \times n_3}$ and $\T{B} \in \mathbb{R}^{n_2 \times n_4 \times n_3}$ is an $n_1 \times n_4 \times n_3$ tensor $\T{C}$ whose $(i,j)$-th tube $\T{C}(i,j,:)$ is given by
	\begin{equation}
	\T{C}(i,j,:) = \sum_{k=1}^{n_2} \T{A}(i,k,:) * \T{B}(k,j,:)
	\end{equation}
\end{defn}
\noindent where $i=1,2,...,n_1$, $j=1,2,...,n_4$. When a third order tensor is viewed as a matrix of tubes along the third dimension, the t-product is analogous to the matrix multiplication except that the multiplication between numbers are replaced by the circular convolution between tubes. 

\begin{remark}
	From the relationship between circular convolution and Discrete Fourier Transform(DFT), the t-product of $\T{A}$ and $\T{B}$ can be computed efficiently in Fourier domain. Specifically, let $\widehat{\T{A}} = \fft(\T{A},[\hspace{2mm}],3)$ and  $\widehat{\T{B}} = \fft(\T{B},[\hspace{2mm}],3)$ be the tensors obtained by taking the Fast Fourier Transform (FFT) along the tube fibers in third dimension of $\T{A}$ and $\T{B}$, then we can compute the t-product of $\T{A}$ and $\T{B}$ through the following,
	\begin{equation}
	\nonumber
	\begin{aligned}
	\widehat{\T{C}}(:,:,i) = &\widehat{\T{A}}(:,:,i)*\widehat{\T{B}}(:,:,i),i=1,2,...,n_3\\
	&\T{C} = \ifft(\widehat{\T{C}} ,[\hspace{2mm}],3)
	\end{aligned}
	\end{equation}
\end{remark}

\begin{defn} \textbf{(Tensor transpose)} The conjugate transpose of a tensor $\T{A} \in \mathbb{R}^{n_1 \times n_2 \times n_3}$ is an $n_2 \times n_1 \times n_3$ tensor $\TA\Tra$ obtained by taking the conjugate transpose of each frontal slice of $\T{A}$, then reversing the order of transposed frontal slices $2$ through $n_3$. 
\end{defn}

\begin{defn}\textbf{(Identity tensor)}
	The identity tensor $\T{I} \in \mathbb{R}^{n \times n \times n_3}$ is defined as follows,
	\begin{equation}
	\T{I}(:,:,1) = I_{n \times n}, \hspace{5mm} \T{I}(:,:,k) = 0, k =2,3,...,n
	\end{equation}
	where $I_{n \times n}$ is the identity matrix of size $n \times n$.
\end{defn}

\begin{defn}\textbf{(Orthogonal Tensor)} A tensor $\T{Q} \in \mathbb{R}^{n \times n \times n_3}$ is orthogonal if it satisfies
	\begin{equation}
	\T{Q}\Tra * \T{Q} = \T{Q}* \T{Q}\Tra = \T{I}
	\end{equation}
\end{defn}

\begin{defn}\textbf{(f-diagonal Tensor)} A tensor is called f-diagonal if each frontal slice of this tensor is a diagonal matrix.
\end{defn}

\subsection{Tensor Singular Value Decomposition(t-SVD)}
We now define the tensor Singular Value Decomposition using the t-product introduced in previous section.
\begin{defn}The t-SVD of a third-order tensor $\T{M} \in \mathbb{R}^{n_1 \times n_2 \times n_3}$ is given by
	\begin{equation}
	\T{M} = \T{U}*\T{S}*\T{V}\Tra
	\end{equation}
	where $*$ denotes the t-product, $\T{U} \in \mathbb{R}^{n_1 \times n_1 \times n_3}$ and $\T{V} \in \mathbb{R}^{n_2 \times n_2 \times n_3}$ are orthogonal tensors. $\T{S} \in \mathbb{R}^{n_1 \times n_2 \times n_3}$ is a rectangular f-diagonal tensor.
\end{defn}

\begin{figure}[htb]
	\centering \makebox[0in]{
		\begin{tabular}{c}
			\includegraphics[height = 0.85in, width = 3.2in]{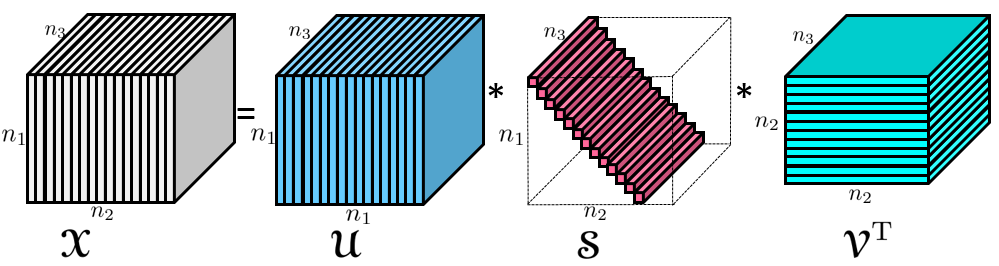}
		\end{tabular}}
		\caption{t-SVD of an $n_1 \times n_2 \times n_3$ tensor.}
		\label{fig:tSVD}
	\end{figure}

Figure~\ref{fig:tSVD} illustrates the t-SVD of $3$rd order tensors. Similar to the t-product, we can also compute t-SVD in Fourier domain, see Algorithm~\ref{alg:tSVD}.

\begin{algorithm}
	\caption{T-SVD of third order tensors}
	\begin{algorithmic}
		\STATE \textbf{Input: } $\T{M} \in \mathbb{R}^{n_1 \times n_2 \times n_3}$
		\STATE \textbf{Output: } $\T{U} \in \mathbb{R}^{n_1 \times n_1 \times n_3}$, $\T{V} \in \mathbb{R}^{n_2 \times n_2 \times n_3}$ and $\T{S} \in \mathbb{R}^{n_1 \times n_2 \times n_3}$ such that $\T{M} = \TU*\T{S}*\TV\Tra$.
		\vspace{1mm}
		\STATE ${\widehat{\T{M}}} = \rm{fft}(\T{M},[\hspace{1mm}],3)$;
		\FOR{$i = 1 \hspace{2mm} \rm{to} \hspace{2mm} n_3$}
		\STATE $ [\M{U}, \M{S}, \M{V}] = \textbf{SVD}(\widehat{\T{M}}(:,:,i))$
		\STATE $ {\widehat{\T{U}}}(:,:,i) = \M{U}; \hspace{1mm} {\widehat{\T{S}}}(:,:,i) = \M{S}; \hspace{1mm} {\widehat{\T{V}}}(:,:,i) = \M{V}; $
		\ENDFOR
		\STATE $\T{U} = \rm{ifft}(\widehat{\T{U}},[\hspace{1mm}],3)$, $\T{S} = \rm{ifft}(\widehat{\T{S}},[\hspace{1mm}],3)$, $\T{V} = \rm{ifft}(\widehat{\T{V}},[\hspace{1mm}],3)$.
	\end{algorithmic}
	\label{alg:tSVD}
\end{algorithm}
	
As discussed in \cite{cvprzemin}, t-SVD has many advantages over the classical tensor decompositions such as CANDECOMP/PARAFAC\cite{parafac1970} and Tucker\cite{Tuck1966c}. For example, given a fixed rank, the computation of CANDECOMP/PARAFAC decomposition can be numerically unstable, since calculating the rank-$1$ components in this model is difficult. Similarly, finding the best Tucker multi-rank $\vec{r}$ approximation to a tensor is numerically expensive and often does not yield the best fit to the original tensor. However, the computation of t-SVD is very easy since one only needs to do several SVDs as shown in Algorithm~\ref{alg:tSVD}. Another very important property is the optimality approximation of t-SVD \cite{Kilmer2011641}, described in the following.
\begin{theorem}
	\label{thm:optimality}
	Let $\T{M} = \TU * \TS *\TV\Tra$ be the t-SVD of $\T{M} \in \mathbb{R}^{n_1 \times n_2 \times n_3}$. Then for $k<\min(n_1,n_2)$, define $\T{M}_k = \sum_{i=1}^{k} \TU(:,i,:) * \TS(i,i,:) *\TV(:,i,:)\Tra$, we have 
	\begin{equation}
	\nonumber
	\T{M}_k = \arg \min_{\tilde{\T{M}} \in \mathbb{M}} \|\T{M} - \tilde{\T{M}}\|_F
	\end{equation}
	where $\mathbb{M} = \{\T{X}*\T{Y} | \T{X} \in \mathbb{R}^{n_1 \times k \times n_3}, \T{Y} \in \mathbb{R}^{k \times n_2 \times n_3}\}$.
\end{theorem}
If we define \emph{\textbf{tensor tubal rank}} of $\T{M}$ to be the number of non-zero diagonal tubes in $\T{S}$\cite{cvprzemin}, then this theorem is saying that $\T{M}_k$ is the closest tensor to $\T{M}$ in Frobenius norm among all tensors of tensor tubal rank at most $k$.

\subsection{t-linear Combination of Tensor Dictionaries and Coefficients}
As in the matrix case, given an overcomplete dictionary $D \in \mathbb{R}^{n \times K}$ which contains $K$ prototype signal-atoms for columns, a signal $y \in \mathbb{R}^{n}$ can be represented as a linear combination of columns of $D$
 \begin{equation}
 \label{eq:linear_combination}
  y=Dx
 \end{equation}
\noindent where $x \in \mathbb{R}^{K}$ is called the representation coefficient vector of $y$. This set up could be easily extended to $3$rd order tensors using the framework outlined in the previous section. Given $K$ tensor columns (or dictionary atoms) $\overrightarrow{\T{D}}_k \in \mathbb{R}^{n_1 \times 1 \times n_3}$, we represent a tensor signal $\overrightarrow{\T{X}} \in \mathbb{R}^{n_1 \times 1 \times n_3}$ using the \emph{\textbf{t-linear combination}} of the given tensor dictionaries as follows,

\begin{equation}
\label{eq:t_linear_combination}
\overrightarrow{\T{X}} = \sum_{k=1}^{K} \overrightarrow{\T{D}}_k * \vec{c}_k = \T{D} * \overrightarrow{\T{C}}
\end{equation}

\noindent where $\{\vec{c}_k\}_{k=1}^K$ are tubes of size $1 \times 1 \times n_3$; $\overrightarrow{\T{C}}\in \mathbb{R}^{K \times 1 \times n_3}$ is called coefficient tensor obtained by aligning all the $\vec{c}_k$. $\T{D} = \{\overrightarrow{\T{D}}_1, \overrightarrow{\T{D}}_2, ..., \overrightarrow{\T{D}}_K\} \in \mathbb{R}^{n_1 \times K \times n_3}$ is the tensor dictionary. The representation (\ref{eq:t_linear_combination}) may either be exact or approximate satisfying
\begin{equation}
 \|\overrightarrow{\T{X}} -  \T{D} * \overrightarrow{\T{C}} \| \le \epsilon
\end{equation}

\noindent for some $\epsilon >0$. When $K>n$, we say the tensor dictionary $\T{D}$ is overcomplete.

\begin{figure}[htb]
	\centering \makebox[0in]{
		\begin{tabular}{c}
			\includegraphics[height = 2in, width = 3.2in]{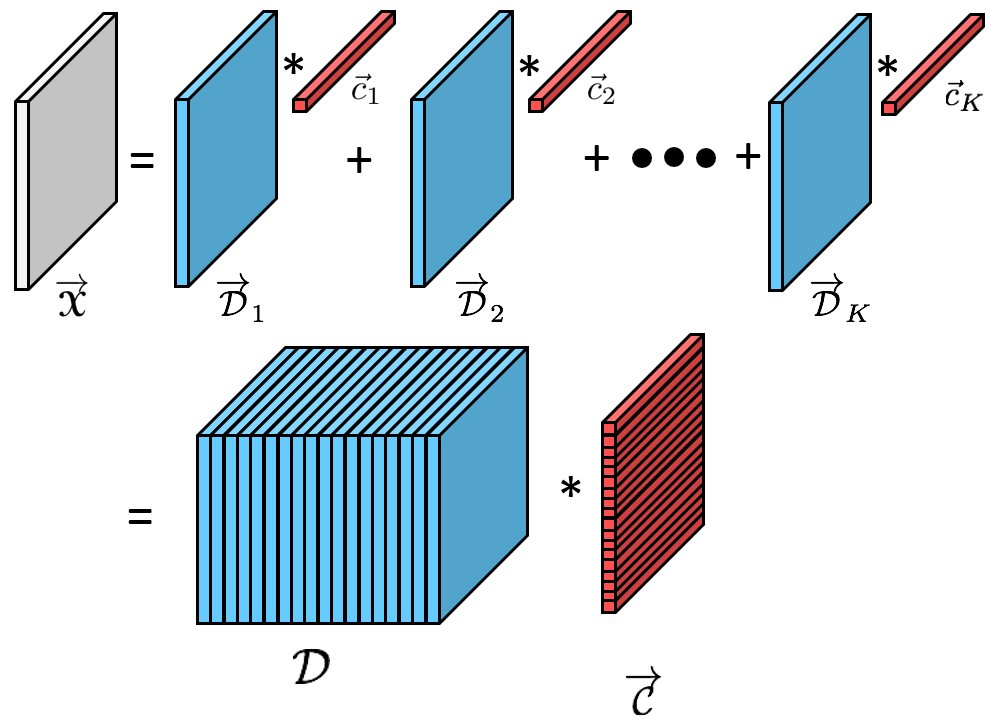}
		\end{tabular}}
		\caption{A tensor signal represented by a t-linear combination of $K$ tensor dictionary atoms.}
		\label{fig:t_linear_combination}
	\end{figure}

\vspace{-2mm}
\section{Problem Formulation}
In this section, we introduce our tensor dictionary learning model and the related algorithm.

\subsection{From Matrix to Tensor Dictionary Learning}
Given an overcomplete dictionary $D \in \mathbb{R}^{n \times K}$ with $K>n$, if $D$ is full rank, there are infinite number of solutions to the representation problem (\ref{eq:linear_combination}); therefore in order to constran the solution set, one common approach is to enforce sparsity. As in classic dictionary learning model which was first designed for the purpose of reconstruction, one adaptively learns an overcomplete dictionary using the training data, which leads to the best possible representation of the data with sparsity constraints. Specifically, given training data $\{y_i\}_{i=1}^{n} \in \mathbb{R}^{d}$ where $d$ is the dimensionality and $n$ is the total number of training data used, dictionary learning methods aim at finding an overcomplete dictionary $D\in \mathbb{R}^{d \times K}$ with $K>d$, and a coefficient matrix $X = [x_1,x_2,...,x_n] \in \mathbb{R}^{K \times n}$ by the following optimization problem,
\begin{equation}
\label{eq:DL}
\begin{aligned}
\min_{D,X} \hspace{2mm} &\sum_{i=1}^{n}\|y_i - Dx_i\|_F^2 \\
\mbox{subject to } \hspace{2mm} &\|x_i\|_q \le T,\hspace{2mm} i=1,2,...,n
\end{aligned}
\end{equation}

\noindent where $\|\cdot\|_q, q\geq 1$ is the $\ell_q$ norm which represents different sparsity regularization. 

Using t-SVD structure discussed in the previous section, we generalize this dictionary learning model to higher dimensional cases. Given training data as tensor columns $\{ {\overrightarrow{\T{Y}}}_i \}_{i=1}^{n} \in \mathbb{R}^{d \times 1 \times n_3}$, we want to find a dictionary $\T{D} \in \mathbb{R}^{n \times K \times n_3}$ with $K>n$, and ``\emph{\textbf{tubal sparse}}" tensor coefficients $\{ \overrightarrow{\T{X}}_i \}_{i=1}^{n} \in \mathbb{R}^{K \times 1 \times n_3}$ to represent the training data using t-product. The tubal sparsity of a tensor column is defined in \cite{cvprzemin} as follows.

\begin{defn} \textbf{(tensor tubal sparsity)} Given a tensor column $\overrightarrow{{\T{X}}}$, the tensor tubal sparsity $\|\cdot\|_\text{TS}$ is defined as the number of non-zero tubes of $\overrightarrow{{\T{X}}}$ in the third dimension.
\end{defn}

\begin{figure}[htb]
	\centering \makebox[0in]{
		\begin{tabular}{c}
			\includegraphics[height = 1.2in, width = 3.2in]{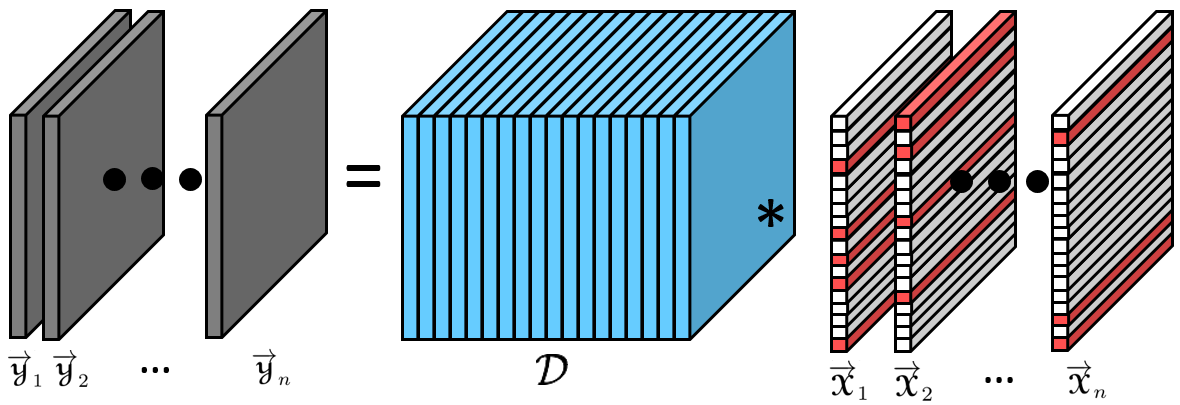}
		\end{tabular}}
		\caption{Data in the form of tensor columns represented by the t-product of tensor dictioanry and tubal-sparse coefficient tensors. The red tubes in the coefficient tensors stand for the non-zero tubes and white ones are zero tubes.}
		\label{fig:tensor_sparse_coding}
	\end{figure}

 Then we can construct our dictionary learning model:
 \begin{equation}
 \label{eq:tensor DL}
 \begin{aligned}
 \min_{\T{D},\overrightarrow{\T{X}}_i} \hspace{2mm} &\sum_{i=1}^{n}\|\overrightarrow{\T{Y}}_i - \T{D}*\overrightarrow{\T{X}}_i\|_F^2  \\
 \mbox{subject to}\hspace{2mm} &\|\overrightarrow{\T{X}}\|_{\text{TS}} \le T,\hspace{2mm}i = 1,2,...,n
 \end{aligned}
 \end{equation}
 
 \noindent or equivalently,
  \begin{equation}
  \label{eq:tensor DL2}
  \begin{aligned}
  \min_{\T{D},\T{X}} \hspace{2mm} &\|\T{Y} - \T{D}*\T{X}\|_F^2  \\
  \mbox{subject to}\hspace{2mm} &\|\T{X}\|_{\text{TS}} \le T_0
  \end{aligned}
  \end{equation}
 
 \noindent where $\T{Y} = \left[\overrightarrow{\T{Y}}_1, \overrightarrow{\T{Y}}_2,...,\overrightarrow{\T{Y}}_n \right] \in \mathbb{R}^{d \times n \times n_3}$ and $\T{X} = \left[\overrightarrow{\T{X}}_1, \overrightarrow{\T{X}}_2,...,\overrightarrow{\T{X}}_n \right] \in \mathbb{R}^{K \times n \times n_3}$. Figure~\ref{fig:tensor_sparse_coding} illustrates the tensor sparse coding model. Note that if the $j$th tube of $\overrightarrow{\T{X}}_i(j,1,:)$ is zero, then it means that the $j$th dictionary $\T{D}(:,j,:)$ is not being used in the representation of $\overrightarrow{\T{Y}}_i$. 
 
\vspace{-1mm}
\subsection{K-TSVD}
We now discuss our tensor dictionary learning model in details. Our model is called K-TSVD since it is a general extension from classic K-SVD to high dimensional tensor based on t-SVD. Similarly to K-SVD algorithm, K-TSVD also consists of two stages: the tensor sparse coding stage and the tensor dictionary update stage. First let's consider the sparse coding stage where the tensor dictionary $\T{D}$ is fixed. So we need to solve
\begin{equation}
\label{eq:sparse_coding}
\begin{aligned}
\min_{\T{X}} \hspace{2mm} &\|\T{Y} - \T{D}*\T{X}\|_F^2  \\
\mbox{subject to}\hspace{2mm} &\|\T{X}\|_{\text{TS}} \le T_0
\end{aligned}
\end{equation}
\noindent or alternatively we can work on an equivalent form,
\begin{equation}
\label{eq:sparse_coding_2}
\min_{\overrightarrow{\T{X}}_i}\|\T{Y} - \T{D}*\T{X}\|_F^2 + \lambda \|\T{X}\|_{\text{TS}}
\end{equation}
\noindent for some positive $\lambda$. Since the sparsity measure is computational intractable in both matrix and tensor cases, we use the $\|\cdot\|_{1,1,2}$ norm \cite{cvprzemin} instead as a convex relaxation for the tubal sparsity, where the $\|\cdot\|_{1,1,2}$ norm of a $3$rd order tensor $\T{X}$ is defined as 
\begin{equation}
\nonumber
\|\T{X}\|_{1,1,2} = \sum_{i,j} \|\T{X}(i,j,:)\|_F
\end{equation}
\noindent If we regard a third dimensional tube $\vec{x} \in \mathbb{R}^{1 \times 1 \times n_3}$ as a $n_3 \times 1$ column vector, then the $\ell_{1,1,2}$ norm of $\T{X}$ is just the summation of $\ell_2$ norm of all such tubes along the third dimension in $\T{X}$.

Replacing the tubal sparsity with the $\ell_{1,1,2}$ norm, the problem becomes
\begin{equation}
\label{eq:sparse_coding_3}
\min_{\T{X}} \|\T{Y} - \T{D}*\T{X}\|^2_F + \lambda \|\T{X}\|_{1,1,2}
\end{equation}
In order to solve this problem, one more definition is needed here. For a third order tensor $\TA \in \mathbb{R}^{n_1 \times n_2 \times n_3}$, define the block diagonal form $\xbar{\T{A}}$ in Fourier domain as follows,
\begin{equation}
\label{eq:blkdiag}
\xbar{\TA} = \text{blkdiag}(\widehat{\TA}) = 
\left[\begin{array}{cccc}\widehat{{\T{A}}}^{(1)}& & & \\
& \widehat{{\T{A}}}^{(2)} & & \\
& &\ddots & \\
& & & \widehat{{\T{A}}}^{(n_3)}\end{array} \right]
\end{equation}

\noindent where $\widehat{\TA} = \fft(\TA,[\hspace{2mm}],3)$ and $\TA^{(i)}$ is the $i$th frontal slice of $\TA$. Then (\ref{eq:sparse_coding_3}) can be equivalently reformulated in Fourier domain as
\begin{equation}
\nonumber
\min_{\xbar{\T{X}}} \|\xbar{\T{Y}} - \xbar{\T{D}} \xbar{\T{X}}\|^2_F + \lambda \sqrt{n_3} \|\widehat{\T{X}}\|_{1,1,2}
\end{equation}
where the $\sqrt{n_3}$ factor comes from the fact that $\|\T{X}\|_F = \|\widehat{\T{X}}\|_F/\sqrt{n_3}$ \cite{cvprzemin}. Use the general framework of Alternating Direction Method of Multipliers (ADMM) \cite{ADMM}, we can solve this optimization problem recursively with the following algorithm:
\begin{align}
\label{eq:admm1}
\xbar{\T{X}}_{k+1} &= \arg\min_{\xbar{\T{X}}} \|\xbar{\T{Y}} - \xbar{\T{D}}\xbar{\T{X}}\|_F^2 + \text{tr}\left( \xbar{\T{Q}}_k\Tra \xbar{\T{X}}\right) + \frac{\rho}{2}\|\xbar{\T{X}} - \xbar{\T{Z}}_k\|_F^2 \\
\label{eq:admm2}
\T{Z}_{k+1} &= \arg\min_{\T{Z}} \|\T{Z}\|_{1,1,2} + \frac{\rho}{2 \lambda} \|\T{X}_{k+1} + \frac{1}{\rho}\T{Q}_k - \T{Z}\|_F^2 \\
\label{eq:admm3}
\T{Q}_{k+1} &= \T{Q}_k + \rho(\T{X}_{k+1} - \T{Z}_{k+1})
\end{align}

\noindent where $\rho>0$. (\ref{eq:admm1}) is essentially a least square minimization problem and we can separately solve it in each frontal slice of $\widehat{\T{X}}$ (or equivalently, each diagonal block of $\xbar{\T{X}}$). Let $\T{C}_{k+1} = \T{X}_{k+1} + \T{Q}_k/\rho$, the update of (\ref{eq:admm2}) is given by

\begin{equation}
\label{eq:sol_admm2}
\begin{aligned}
\T{Z}_{k+1}(i,j,:) = &\left( 1-\frac{\lambda}{\rho \|\T{C}_k(i,j,:)\|_F} \right)_+ \T{C}(i,j,:) \\
& \forall i=1,2,...,K,\hspace{1mm}j = 1,2,...,n
\end{aligned}
\end{equation}

\noindent where $(\cdot)_+ = \max (0,\cdot) $.
\\

The second stage of our tensor dictionary learning model is dictionary update. Given fixed $\T{D}$ and $\T{X}$, suppose we only want to update the $k$-th element of $\T{D}$, we can decompose the error term as follows,
\begin{equation}
\nonumber
\begin{aligned}
&\|\T{Y} - \T{D}*\T{X}\|^2_F \\
= &\left\| \T{Y} - \sum_{j=1}^{K} \overrightarrow{\T{D}}_j*\T{X}(j,:,:) \right\|_F^2 \\
= &\left\| \left(\T{Y} - \sum_{j \neq k}\overrightarrow{\T{D}}_j*\T{X}(j,:,:)\right) -  \overrightarrow{\T{D}}_k*\T{X}(k,:,:)\right\|_F^2 \\
= &\|\T{E}_k - \overrightarrow{\T{D}}_k*\T{X}(k,:,:) \|_F^2 \\
= &\|\T{E}_k - \T{D}(:,k,:)*\T{X}(k,:,:) \|_F^2
\end{aligned}
\end{equation}
\noindent $\T{E}_k$ here stands for the representation error when the $k$-th atom $\T{D}(:,k,:)$ is removed from the dictionary. The next step is to find $\T{D}(:,k,:) * \T{X}(k,:,:)$ which best approximates $\T{E}_k$, so that the error term is minimized. This is essentially to compute the best tubal rank-$1$ approximation using Theorem~\ref{thm:optimality}. Since we need to maintain the tubal sparsity of $\T{X}$ and don't want to fully fill $\T{X}(k,:,:)$, let $w_k = \{i| \T{X}(k,i,:) \neq 0,i=1,2,...,n\}$ be the set of indices where data $\T{Y}$ uses tensor dictionary $\T{D}(:,k,:)$ and restrict $\T{E}_k$ by choosing the tensor columns corresponding to $w_k$ to obtain $\T{R}_k: \T{R}(:,i,:) = \T{E}(:,w_k(i),:), i= 1,2,...,|w_k|$. From Theorem~\ref{thm:optimality}, we apply t-SVD on $\T{R}_k$ to get $\TU,\TS$ and $\TV$, and take the first tensor column of $\TU$ to update $\T{D}(:,k,:)$, use $\T{S}(1,1,:)*\TV(:,1,:)\Tra$ to renovate the coefficient tensors which use the $k$-th dictionary. To accelerate the algorithm we only compute the approximate rank-1 SVDs in Fourier domain when we compute t-SVD of $\T{R}$. The complete algorithm is presented in Algorithm~\ref{alg:k-tsvd}.

\begin{algorithm} [thb]
	\caption{K-TSVD}
	\label{alg:k-tsvd}
	\textbf{Input }: Observed tensor data $\T{Y} = \{\overrightarrow{\T{Y}}_i\}_{i=1}^{n_2} \in \mathbb{R}^{n_1 \times n_2 \times n_3}$, $\lambda >0$.\\
	\textbf{Initialize}: Dictionary $\T{D}_0 \in \mathbb{R}^{n_1 \times K \times n_3}$ \\
	\textbf{Repeat until convergence}:\\
	\vspace{-5mm}
	\begin{algorithmic}[1]
		\STATE Compute the sparse coefficient tensor using (\ref{eq:admm1})-(\ref{eq:admm3}):
		\begin{equation}
		\nonumber
		\T{X} = \arg\min_{\T{X}} \hspace{2mm}\|\T{Y} - \T{D}*\T{X}\|_F^2 + \lambda \|\T{X}\|_{1,1,2}
		\end{equation}
		\FOR{$k=1,2,...,K$}
		    \STATE Let $w_k = \{i| \T{X}(k,i,:) \neq 0\}$ be the set of indices where data $\T{Y}$ uses dictionary $\T{D}(:,k,:)$.
		    \STATE Compute $\T{E}_k = \T{Y} - \sum_{j \neq k} \T{D}(:,j,:)*\T{X}(j,:,:)\Tra$, which is the over all error without using the $k$-th dictionary atom $\T{D}(:,k,:)$.
		    \STATE Restrict $\T{E}_k$ by choosing only the tensor columns corresponding to $w_k$ and obtain $\T{R}_k$: 
		    \begin{equation}
		    \T{R}(:,i,:) = \T{E}(:,w_k(i),:)
		    \end{equation}
		    for $i= 1,2,...,|w_k|$.
		    \STATE Compute the t-SVD of $\T{R}_k$:
		    \begin{equation}
		    \nonumber
		    \T{R}_k = \TU * \TS * \TV\Tra.
		    \end{equation}
		    \vspace{-5mm}
		    \STATE Update $\T{D}(:,k,:) = \T{U}(:,1,:)$.
		    \STATE Update $\T{X}(k,w_k,:) = \T{S}(1,1,:)*\TV(:,1,:)\Tra$.
			\ENDFOR
		\end{algorithmic}
		\textbf{Output}: Trained tensor dictionary $\T{D}$.\\
	\end{algorithm}
\section{Experiment Results}

\subsection{Filling Missing Pixels in Tensors}
In this section we consider the application of filling missing pixels in third order tensors. Suppose that we are given a video with dead pixels, where the dead pixels mean pixel values are deleted or missing on some fixed positions of each frame. Specifically, let $\Omega$ indicate the set of indices of the remaining pixels and $\T{M}$ be the data tensor, then $\T{M}(i,j,:) = 0$ for all $(i,j) \notin \Omega$.
Our goal is to recover such tensors with missing pixels. Suppose $\T{D}$ is the learned overcomplete dictionary on the training data, define $P_\Omega$ as an orthogonal projector such that $P_\Omega(\T{M})(i,j,:) = \T{M}(i,j,:)$, if $(i,j) \in \Omega$ and $0$ otherwise. Then for each patch $\overrightarrow{\T{M}}_k$ in the test data, the reconstruction of this patch is $\T{D}*\overrightarrow{\T{C}}_k$, where $\overrightarrow{{\T{C}}}_k$ is the solution to
\begin{equation}
\label{eq:complete}
\min_{\overrightarrow{\T{C}}_k} \|P_\Omega(\overrightarrow{{\T{M}}}_k) - P_\Omega(\T{D}*\overrightarrow{\T{C}}_k )\|_F^2 + \lambda \|\overrightarrow{\T{C}}_k\|_{1,1,2}
\end{equation} 
\noindent which can be solved in the same manner as (\ref{eq:sparse_coding_3}).

We utilized a basketball video here to apply K-TSVD algorithm and reconstruct $\T{M}$ from missing pixels. There are $40$ frames in the video and the resolution of each frame is $144 \times 256$. To learn the overcomplete dictionary using K-TSVD, we randomly took $9000$ overlapping block patches of size $8 \times 8 \times 10$ from the first $30$ frames, saved them as tensor columns of size $64 \times 1 \times 10$, and obtained our training data $\T{Y}$ of total size $64 \times 9000 \times 10$. All these patches were used to train a tensor dictionary with $K=256$ atoms. The last $10$ frames of the video were used for testing. We took the total $576$ disjoint $8 \times 8 \times 10$ blocks in the last $10$ frames, saved each block into a tensor column, and obtained our training data of size $64 \times 576 \times 10$.

\begin{figure}[htbp]
	\centering \makebox[0in]{
		\begin{tabular}{c c}
			\includegraphics[height = 1.4in, width = 1.6in]{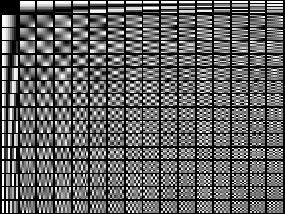}      
			\includegraphics[height = 1.4in, width = 1.6in]{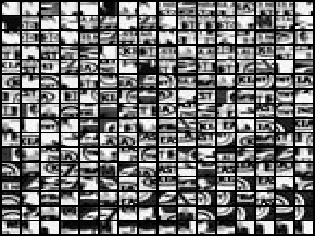}\\
			\includegraphics[height = 1.4in, width = 1.6in]{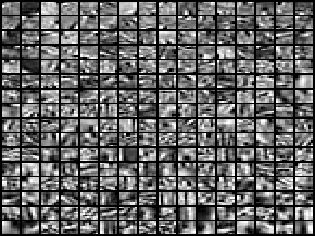}      
			\includegraphics[height = 1.4in, width = 1.6in]{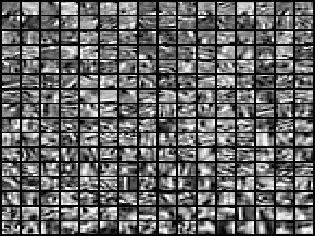} 
		\end{tabular}}
		\caption{\textbf{Upper left}: The overcomplete DCT dictionary. \textbf{Upper right}: Dictionary learned on the first frame of the basketball video using K-SVD. \textbf{Lower left} The first frontal slice $\T{D}(:,:,1)$ of the learned dictionary of the tensor. \textbf{Lower right} The $3$rd frontal slice $\T{D}(:,:,3)$ of the learned dictionary of the tensor.}
		\label{fig:dictionary_basketball}
	\end{figure}

\begin{figure}[htbp]
	\centering \makebox[0in]{
		\begin{tabular}{c c}
			\includegraphics[width=.49\linewidth]{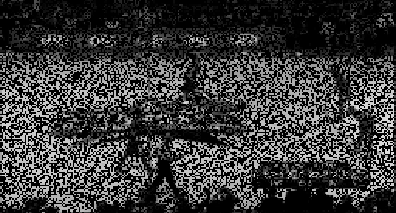}
			\includegraphics[width=.49\linewidth]{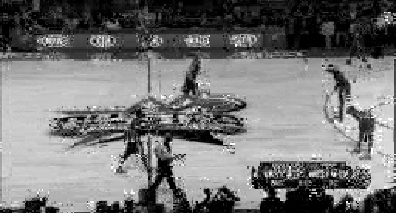}\\
			\includegraphics[width=.49\linewidth]{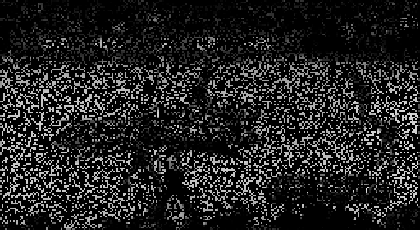}
			\includegraphics[width=.49\linewidth]{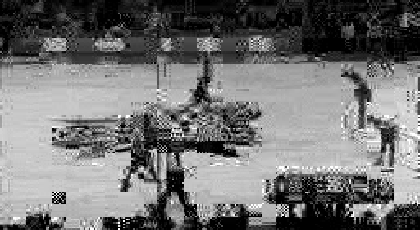}\\
			\noindent\footnotesize{(a) Frame with missing pixels~~~~~~~~~~~~~~~
				(b) DCT reconstruction~~~~~~~~}\\
			\includegraphics[width=.49\linewidth]{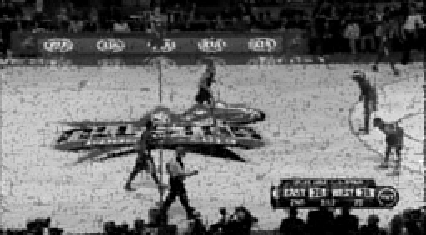}
			\includegraphics[width=.49\linewidth]{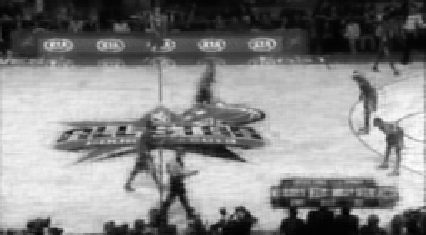}\\
			\includegraphics[width=.49\linewidth]{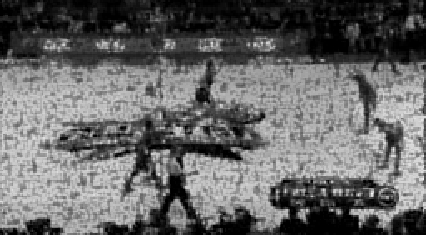}
			\includegraphics[width=.49\linewidth]{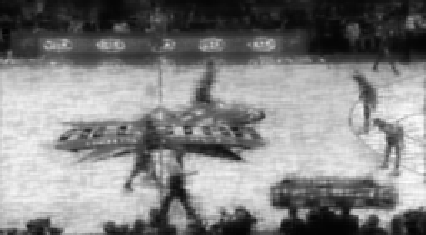}\\
			\noindent\footnotesize{(c) K-SVD reconstruction ~~~~~~~~~~~~~~~
				(d) K-TSVD reconstruction~~}\\
		\end{tabular}}
		\caption{The reconstruction result from missing pixels on the basketball video. The different rows are for $50\%$ and $70\%$ of missing pixels respectively.}
		\label{fig:fill_missing_basketball}
	\end{figure}

\begin{figure}
		\centering 
			\includegraphics[height = 2in, width = 3in]{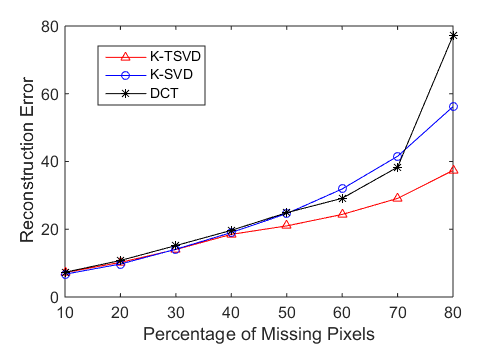}    
			\caption{The reconstruction error comparison of DCT, K-SVD and K-TSVD on the reconstruction. The sparsity varies from $10\%$ to $80\%$. }
			\label{fig:fill_missing_compare}
		\end{figure}

We investigated the performance of K-TSVD by comparing it with K-SVD and DCT. In K-SVD, in order to have a fair comparison, for each test frame we also randomly trained $10000$ block patches of size $ 8 \times 8$ in the first $30$ frames. We visualize an example of the overcomplete DCT dictioanry, the K-SVD learned dictionary and the K-TSVD learned dictionary in Figure~\ref{fig:dictionary_basketball}. One frame with $50\%$ and $70\%$ missing pixels and its reconstructions are shown in Figure~\ref{fig:fill_missing_basketball}. As one can see the reconstruction based on K-TSVD learned dictionary has a better quality. Figure~\ref{fig:fill_missing_compare} shows the reconstruction error (RE) comparison of those three approaches, where the error is computed via $\text{RE} = \sqrt{\|\T{X} - \T{X}_{\text{rec}}\|_F^2/N}$, $N$ is the total number of pixels in the data. We can see that when the percentage of missing pixels is small, all three methods perform equally well. With more missing pixels, K-TSVD gives better performance over the other two methods.

\vspace{-4mm}
\subsection{Multispectral Image and Video Denoising}
In order to further test the proposed method, we applied our algorithm on multispectral/hyperspectral images and video data denoising. In the first experiment the multispectral data was from the \textbf{Columbia datasets} \footnote{\url{http://www1.cs.columbia.edu/CAVE/databases/multispectral/}}, each dataset contains 31 real-world images of size $512 \times 512$ and is collected from $400$nm to $700$nm at $10nm$ steps. In our experiment we resized each image into size of $205 \times 205$, and took images of the last $10$ bands to accelerate the speed of training tensor dictionaries. Therefore the total size of the tensor data we used here is $205 \times 205 \times 10$. Further work is required to fully deploy the algorithm in large-scale high order tensor applications. 

For the noise model we consider the fixed-location defects without knowing the noisy positions, which commonly exists in video and multispectral images. On image of each bandwidth, some fixed pixel locations are corrupted with very high noise and our task is to recover the image. Specifically in our experiment we picked a sparse number of pixel locations and added Gaussian noise on these positions of each image. Let $\Omega$ indicate the set of noisy pixel locations, then what we did was for each $(i,j) \in \Omega$, $k=1,2,...,10$, $\T{Y}(i,j,k) = \T{Y}(i,j,k)+w_{ijk}$, where $\T{Y}$ is the clean tensor and $w_{ijk} \sim \mathcal{N}(0,\sigma)$ is the additive Gaussian noise.
 
\begin{figure}[htbp]
	\centering \makebox[0in]{
		\begin{tabular}{c c}   
			\includegraphics[height = 1.4in, width = 1.6in]{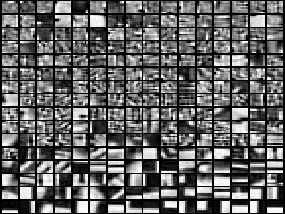}
			\includegraphics[height = 1.4in, width = 1.6in]{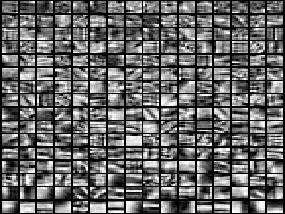}      
		\end{tabular}}
		\caption{\textbf{Left} The learned dictionary on the first image using K-SVD. \textbf{Right} The first frontal slice $\T{D}(:,:,1)$ of the learned dictionary of the tensor.}
		\label{fig:dictionary_toy}
	\end{figure}

To train the data and learn the dictionaries, similarly to what we did in the previous experiment, we randomly took $10000$ overlapping patches of size $8\times 8 \times 10$ from the noisy tensor data, and saved each patch into a tensor column of size $64 \times 1 \times 10$. Therefore the tensor $\T{Y}$ to train here was of size $64 \times 10000 \times 10$. Since the total number of overlapping patches is $(205-7)^2 = 39204$, we only trained about a quarter of all the overlapping patches for the reason of computation time. If the size of data gets larger, then more patches are needed to ensure a more accurate dictionary. For a fair comparison, in K-SVD we also randomly select $10000$ overlapping patches of size $8 \times 8$ within each noisy image. The trained dictionaries of KSVD and K-TSVD on the noisy tensor data are shown in Figure~\ref{fig:dictionary_toy}.

The denoising process of our method includes a tensor sparse coding stage based on the learned tensor dictionary. We extracted each $8 \times 8 \times 10$ patch in the noisy multispectral images and solved the tensor sparse coding problem (\ref{eq:sparse_coding_3}) to obtain the denoised patch. Following a similar idea in \cite{ksvddenoise}, we averaged all the denoised patches with some relaxation obtained by averaging with the original noisy data then got our denoised tensor. 

\begin{figure}[htbp]
	\centering \makebox[0in]{
		\begin{tabular}{c c}
			\includegraphics[width=.33\linewidth]{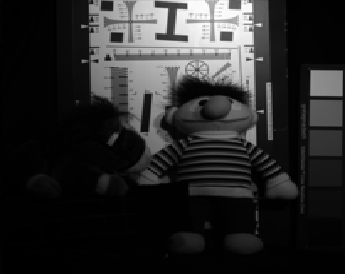}
			\includegraphics[width=.33\linewidth]{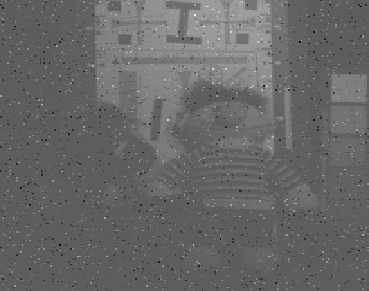}
			\includegraphics[width=.33\linewidth]{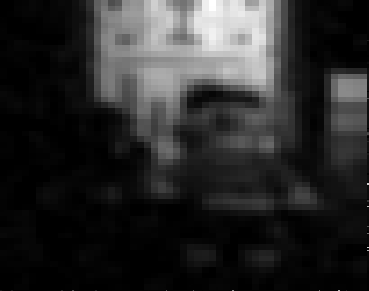}\\
			\noindent\footnotesize{~~~~~(a) Clearn image ~~~~~~~~~~~
				(b) Noisy image ~~~~~~~~
				(c) Bandwise K-SVD ~~}\\
			\includegraphics[width=.33\linewidth]{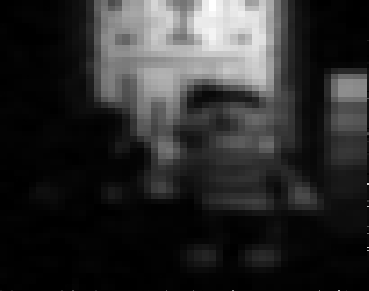}
			\includegraphics[width=.33\linewidth]{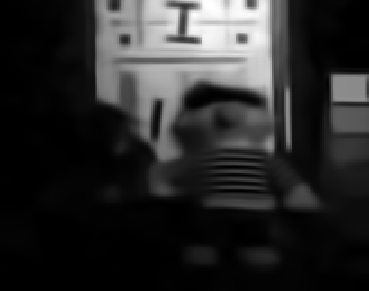}
			\includegraphics[width=.33\linewidth]{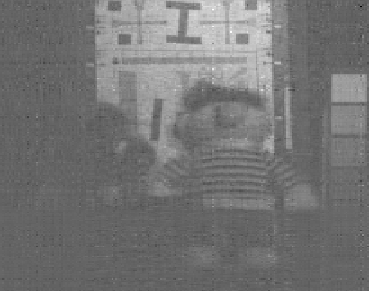}\\
			\noindent\footnotesize{(d) 3DK-SVD ~~~~~~~~~~
				(e) Bandwise BM3D ~~~~~~~~~~~~~~~
				(f) LRTA ~~~~}\\
			\includegraphics[width=.33\linewidth]{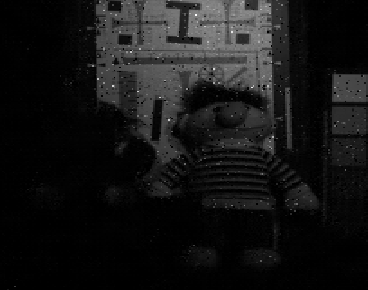}
			\includegraphics[width=.33\linewidth]{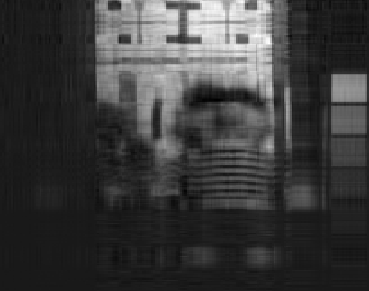}
			\includegraphics[width=.33\linewidth]{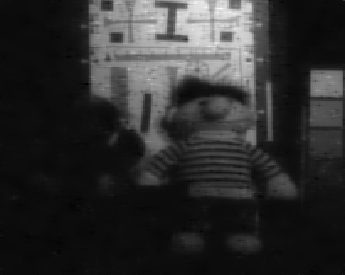}\\
			\noindent\footnotesize{~~(g) DNMDL ~~~~~~~~~~~~~~~~~
				(h) PARAFAC ~~~~~~~~~~~~~~~~~
				(i) \textbf{K-TSVD} ~~}\\
		\end{tabular}}
		\caption{Denoised image at the $610$nm band of chart and stuffed toy. The sparsity of the noisy pixels is $10$\% and the locations of noisy pixels are consistent on image of each band. The additive noise is Gaussian with $\sigma = 100$.}
		\label{fig:denoise_toy}
	\end{figure}

To test the performance of our method, we compared our K-TSVD to these methods: K-SVD (band-wise)\cite{ksvd,ksvddenoise} 3D K-SVD \cite{ksvddenoise}, BM3D (band-wise) \cite{bm3d}, LRTA\cite{LRTA}, DNMDL\cite{tucker2} and PARAFAC\cite{parafac}. BM3D is a non-local denoising method based on an enhanced sparse representation in the transform domain, achieved by grouping similar patches into 3D data arrays. DNMDL is a Tucker dictionary learning based method, which like BM3D first groups the 3D patches and then use Tucker dictionary learning approach within each group to denoise. These two methods take the non-local similarity properties of different patches into consideration, and have very good denoising performance on some cases. LRTA is a Tucker3 based method which simply employs a low rank tensor approximation in Tucker3 model as denoised images. Similarly, PARAFAC is a CANDECOMP/PARAFAC based approach and it also obtains denoising result using a low CP rank approximation. Therefore these two methods can be regarded as a same type of denoising approach. K-SVD, 3DK-SVD and our method K-TSVD perform denoising by learning a overcomplete dictionary on the noisy data and reconstruct the image using sparse coding, which is different from the other methods. The result with $\sigma = 100$ and the sparsity of noisy pixels equaling $10\%$ is shown in Figure~\ref{fig:denoise_toy}. The detailed PSNR comparison on different noise levels of these methods is in Table~\ref{tab:toy_compare}. We can see that our algorithm has a better performance over the other competing methods on most cases.

\begin{table}
	\begin{center}
		\caption{PSNR(dB) of chart and stuffed toy images.}
		\begin{tabular}{|c|c|c|c|c|c|} \hline
		    Sparsity& $5\%$ & $10\%$ & $15\%$  & $10\%$ & $10\%$  \\ \space
			Noise level& $100$ & $100$ & $100$  & $150$ & $200$  \\ \hline
			Noisy image& 20.96 & 18.18 & 16.35  &14.75 &12.10 \\ \hline
			K-SVD &22.73& 22.60 & 22.49  &22.38  &22.00 \\ \hline
			3DK-SVD &22.61 &22.53&22.47    &22.41  &22.20  \\ \hline
			BM3D & 26.95& 26.62 & 26.36    &25.23  &24.29 \\ \hline
			LRTA & 23.54 & 26.84 & 26.65   &23.90  &22.03 \\ \hline
			DNMDL & 24.07 & 23.73& 25.16   &17.89  &16.83  \\ \hline
			PARAFAC & 27.07& 26.86 & 26.72 &26.13 & 25.24\\ \hline
			\textbf{KTSVD} & \textbf{27.19}  & \textbf{26.98} & \textbf{26.79}   &\textbf{26.18} & \textbf{25.44} \\ \hline
		\end{tabular}
		\label{tab:toy_compare}
	\end{center}
\end{table}

\begin{figure}[htbp]
	\centering \makebox[0in]{
		\begin{tabular}{c c}
			\includegraphics[width=.33\linewidth]{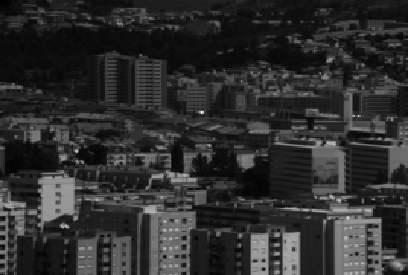}
			\includegraphics[width=.33\linewidth]{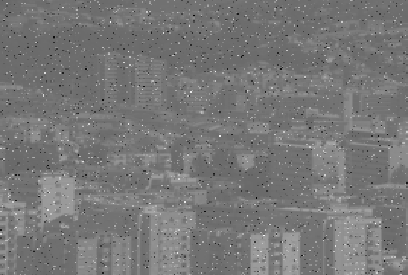}
			\includegraphics[width=.33\linewidth]{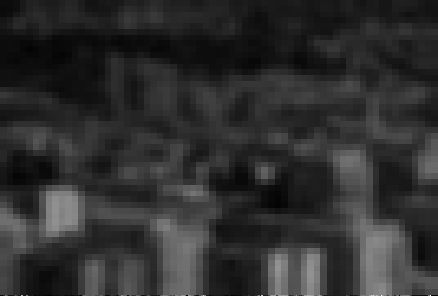}\\
			\noindent\footnotesize{~~~~~(a) Clearn image ~~~~~~~~~~~
				(b) Noisy image ~~~~~~~~
				(c) Bandwise K-SVD ~~}\\
			\includegraphics[width=.33\linewidth]{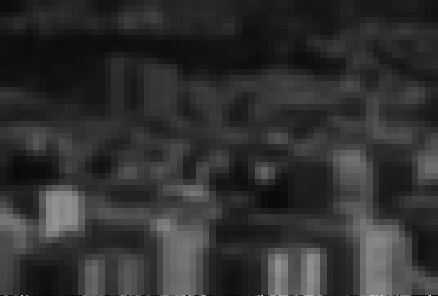}
			\includegraphics[width=.33\linewidth]{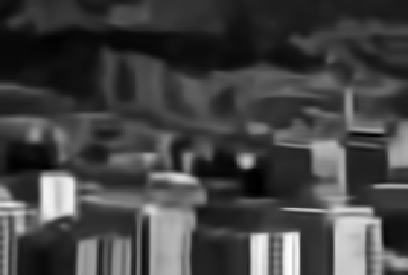}
			\includegraphics[width=.33\linewidth]{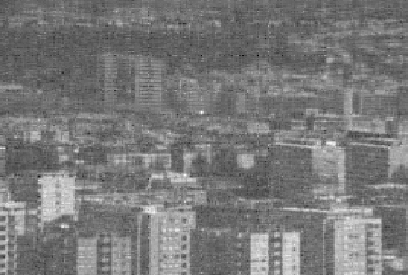}\\
			\noindent\footnotesize{(d) 3DK-SVD ~~~~~~~~~~
				(e) Bandwise BM3D ~~~~~~~~~~~~~~~
				(f) LRTA ~~~~}\\
			\includegraphics[width=.33\linewidth]{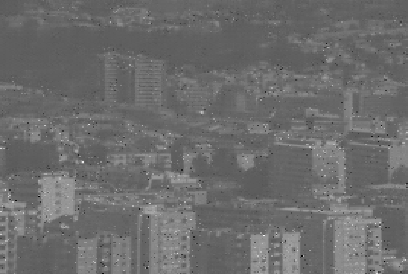}
			\includegraphics[width=.33\linewidth]{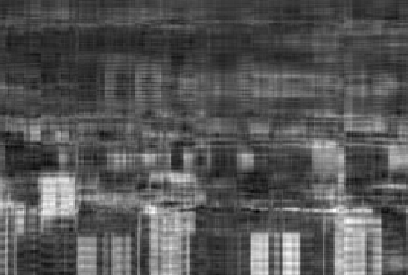}
			\includegraphics[width=.33\linewidth]{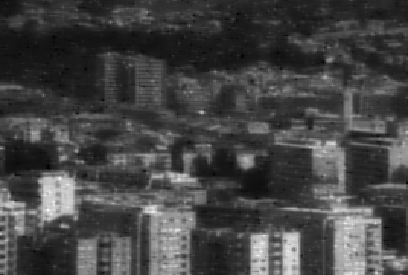}\\
			\noindent\footnotesize{~~(g) DNMDL ~~~~~~~~~~~~~~~~~
				(h) PARAFAC ~~~~~~~~~~~~~~~~~
				(i) \textbf{K-TSVD} ~~}\\
		\end{tabular}}
		\caption{Denoised image at the $700$nm band of hyperspectral images on natural scene. The sparsity of the noisy pixels is $10$\% and the locations of noisy pixels are consistent on image of each band. The additive noise is Gaussian with $\sigma = 100$.}
		\label{fig:denoise_scene}
	\end{figure}

The second dataset we used was a set of hyperspectral images of natural scenes \cite{hyperspectral}. Similarly as before, we only took the images from bandwidth $630$nm to $720$nm and obtain a clean tensor of size $205 \times 268 \times 10$. We trained 10000 dictionaries on the noisy data and perform denoising process using the same technique. The performance is shown in Figure~\ref{fig:denoise_scene} and the PSNR comparison on different noise levels is given in Table~\ref{tab:scene_compare}. In this dataset, the PSNR result shows that our algorithm also gives the good denoising performance on most cases. As one of the tensor based approach, LRTA gives the best PSNR on the case of sparsity $10\%$ and standard deviation of the noise being $100$. PARAFAC also works pretty well when sparsity equals $10\%$ and noise level is $200$.
\vspace{-1mm}

\begin{figure}[htbp]
	\centering \makebox[0in]{
		\begin{tabular}{c c}
			\includegraphics[width=.33\linewidth]{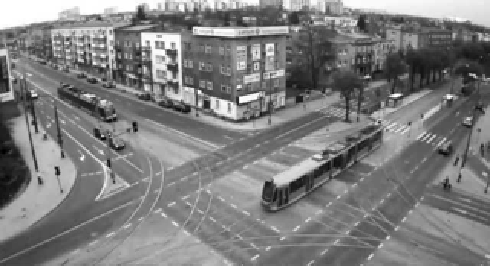}
			\includegraphics[width=.33\linewidth]{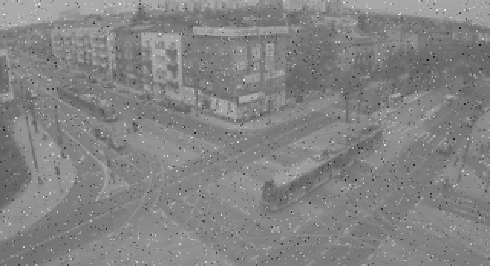}
			\includegraphics[width=.33\linewidth]{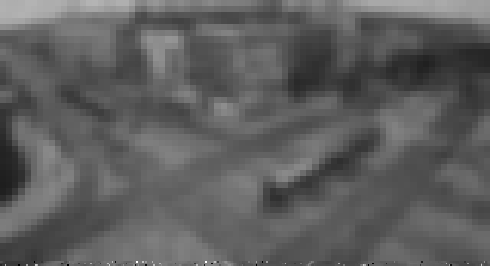}\\
			\noindent\footnotesize{~~~~~(a) Clearn image ~~~~~~~~~~~
				(b) Noisy image ~~~~~~~~
				(c) Bandwise K-SVD ~~}\\
			\includegraphics[width=.33\linewidth]{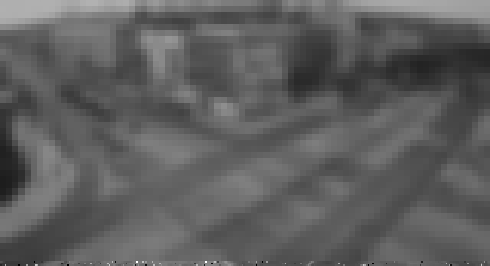}
			\includegraphics[width=.33\linewidth]{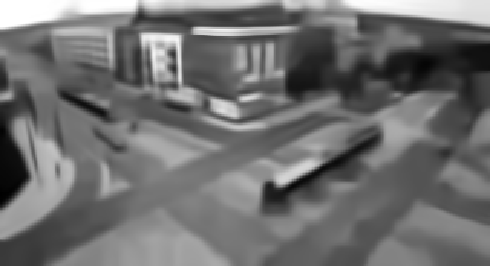}
			\includegraphics[width=.33\linewidth]{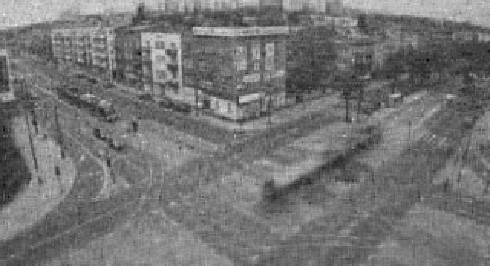}\\
			\noindent\footnotesize{(d) 3DK-SVD ~~~~~~~~~~
				(e) Bandwise BM3D ~~~~~~~~~~~~~~~
				(f) LRTA ~~~~}\\
			\includegraphics[width=.33\linewidth]{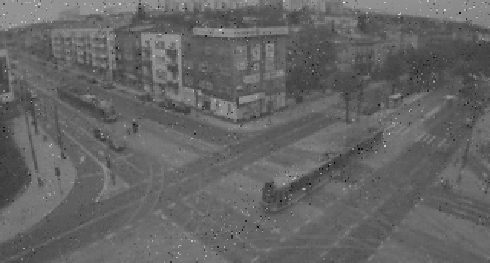}
			\includegraphics[width=.33\linewidth]{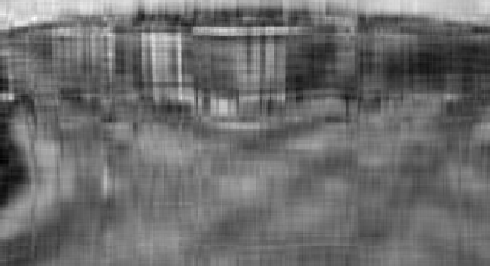}
			\includegraphics[width=.33\linewidth]{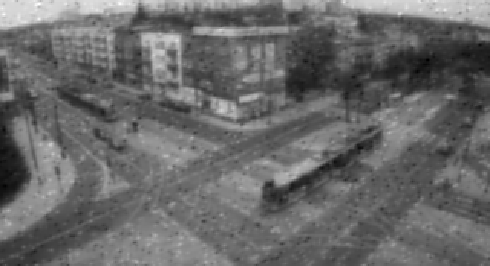}\\
			\noindent\footnotesize{~~(g) DNMDL ~~~~~~~~~~~~~~~~~
				(h) PARAFAC ~~~~~~~~~~~~~~~~~
				(i) \textbf{K-TSVD} ~~}\\
		\end{tabular}}
		\caption{Video denoising result. The sparsity is $10\%$ and $\sigma = 100$.}
		\label{fig:denoise_video}
	\end{figure}

\begin{table}
	\begin{center}
		\caption{PSNR(dB) of natural scene images.}
		\begin{tabular}{|c|c|c|c|c|c|} \hline
			Sparsity& $5\%$ & $10\%$ & $15\%$  & $10\%$ & $10\%$  \\ \space
			Noise level& $100$ & $100$ & $100$  & $150$ & $200$  \\ \hline
			Noisy image& 21.29 & 18.02 & 16.45  &14.62 &12.19 \\ \hline
			K-SVD &22.81& 22.70 & 22.64  &22.51  &22.28 \\ \hline
			3DK-SVD &22.78 &22.73&22.71    &22.66  &22.58  \\ \hline
			BM3D & 24.93& 24.56 & 24.37    &23.56  &22.90 \\ \hline
			LRTA & 25.64 & 25.68 & \textbf{26.12}    &23.76  &21.96 \\ \hline
			DNMDL & 22.01 & 23.40& 24.62   &20.68  &18.47  \\ \hline
			PARAFAC & 24.57& 24.48 & 24.39 &24.21 & \textbf{23.60} \\ \hline
			\textbf{KTSVD} & \textbf{25.94}  & \textbf{25.73} & 25.53 &\textbf{24.96} &23.55 \\ \hline
		\end{tabular}
		\label{tab:scene_compare}
	\end{center}
\end{table}
We also applied K-TSVD algorithm on video denoising. The video that we used here was footage from a still camera view of a traffic intersection \footnote{\url{www.changedetection.net}}. The resolution of each frame is $175 \times 328$, and we performed our method on every $10$ frames. Figure~\ref{fig:denoise_video} shows one frame of the denoising result with sparsity $=10\%$ and noise level $100$. As one can see in this experiment both LRTA and K-TSVD perform well.
\vspace{-2mm}
\vspace{-5mm}
\vspace{-1mm}
\section{Conclusion}
\vspace{-0.5mm}
In this paper, we present a new method for tensor dictionary learning algorithm K-TSVD, using the t-SVD framework. Our main contribution lies in explicitly integrating the sparse coding of third order tensors in t-SVD sense, and based on this we generalize the K-SVD dictionary learning model to deal with higher order tensors. The experimental results show that our approach yields very good performance on video completion and multispectral images denoising. Possible future work includes applying the group technique used in BM3D and DNMDL to process groups of similar patches separately.

{\small
\bibliographystyle{ieee}
\bibliography{egbib}
}

\end{document}